\documentclass[a4paper]{article}

\usepackage{INTERSPEECH2022}
\usepackage{array}
\usepackage{multicol}
\usepackage{tabularx}

\newcolumntype{Y}{>{\raggedright\arraybackslash}X}
\newcolumntype{L}[1]{>{\raggedright\arraybackslash}p{#1}}
\newcolumntype{C}[1]{>{\centering\arraybackslash}p{#1}}
\newsavebox{\resulttables}
\makeatletter
\newcommand{\nextleftpage}{%
  \if@firstcolumn
    \newpage\null\newpage
  \else
    \newpage
  \fi}
\makeatother

\title{The Bairong System for MLC-SLM 2026: Dynamic Question-Aware Evidence Routing for Multilingual Conversational Speech Understanding}
\name{Shangkun Huang, Junchao Hu, Huan Shen, Guoji Wang, Yingao Wang, Shaosai Li, Wei Zou, Yunzhang Chen}
\address{BRVoice Team, Bairong, Inc., China}
\email{\{shangkun.huang,junchao.hu,huan.shen\}@brgroup.com}

\begin{document}

\maketitle

\begin{abstract}
Long multilingual conversational spoken question answering requires systems to balance long-range transcript semantics with sparse acoustic and speaker-sensitive cues. We present the Bairong system for the MLC-SLM 2026 Challenge, where a diarization-ASR front-end produces speaker-attributed transcripts and a dynamic evidence router constructs question-specific inputs for answer prediction. Instead of applying a fixed transcript-only or audio-only policy, the router infers the required evidence type and context scope from the question and answer options, and selects among full transcript context, local audio-text fusion, speaker-linked evidence, and compact global acoustic samples. This transcript-backbone design keeps discourse context available while activating audio only when it provides complementary evidence. Our Task 1 system achieves 25.70\% and 18.44\% tcpMER on the development and evaluation sets. For Task 2, the final system obtains 94.84\% development accuracy, outperforming the full-transcript baseline by 1.68 points and the best audio-centric diagnostic system by 2.77 points. These results support dynamic question-aware routing as an effective evidence allocation strategy for conversational spoken QA.
\end{abstract}

\section{Introduction}

Long conversational spoken question answering (SQA) is not merely an ASR problem or a generic audio-understanding problem~\cite{lee2018spoken,wu2023heysquad,zhao2024librisqa,shankar2024coraalqa,wang2025mmsu}. In multilingual two-speaker conversations, different questions depend on different evidence forms: some require long-range discourse semantics, some require who-said-what speaker attribution, and others require prosody, laughter, pace, emphasis, or acoustically ambiguous words~\cite{shen2026codett}. The MLC-SLM Challenge targets this setting under a no-oracle inference condition, where systems must perform diarization, recognition, and question answering without oracle segmentation or speaker labels~\cite{mlcslm2026,mu2025summary}.

Existing MLC-SLM systems reflect two complementary directions. End-to-end speech LLMs can model diarization and recognition jointly for long multilingual conversations~\cite{saengthong2025unified,peng2026vibevoice}, while context-aware and cascaded systems exploit multimodal history or speaker-attributed transcripts~\cite{mu2026hearing,mu2025summary,xue2025teaaslp}. Recent models such as Qwen3-Omni, Qwen3.5-Omni, Qwen3-ASR, and Omni-Captioner further strengthen acoustic, textual, and fine-grained multimodal modeling~\cite{xu2025qwen3omni,team2026qwen3,shi2026qwen3asr,ma2025omni}. However, these advances do not remove the evidence-selection problem. Transcript-only input preserves long-range semantics but loses prosody, laughter, speaker timbre, and information needed to resolve ASR ambiguity~\cite{huang2025leveraging}. Audio-only input preserves those cues, but long recordings increase computation, introduce irrelevant regions, and can dilute the answer model's attention.

The central question is therefore not whether audio or transcripts are universally better, but which evidence should be exposed for each question. A fixed input policy cannot simultaneously preserve global discourse context and isolate sparse acoustic cues. The resulting research gap is question-conditioned evidence selection: jointly deciding what evidence type is needed and whether that evidence is local or global.

We address this gap with a hybrid question-aware evidence routing framework. It combines LLM-based coarse evidence estimation with deterministic anchor and speaker-cue parsing to construct question-specific transcript, local audio-text, speaker-linked, or global acoustic evidence.

Our contributions are threefold. First, we propose this hybrid router, integrating coarse LLM estimates with deterministic anchor and speaker-cue parsing. Second, we instantiate it in a transcript-backbone MLC-SLM Task 2 system that allocates semantic, acoustic, and speaker-linked evidence per question. Third, transcript-backed ablations and audio-centric diagnostics show that hybrid routing outperforms static transcript routing, full-transcript input, and audio-centric alternatives; question-conditioned allocation matters more than simply using more audio.

\section{Proposed System}

\subsection{Overall Pipeline}

\begin{figure*}[t]
  \centering
  \includegraphics[width=0.98\textwidth]{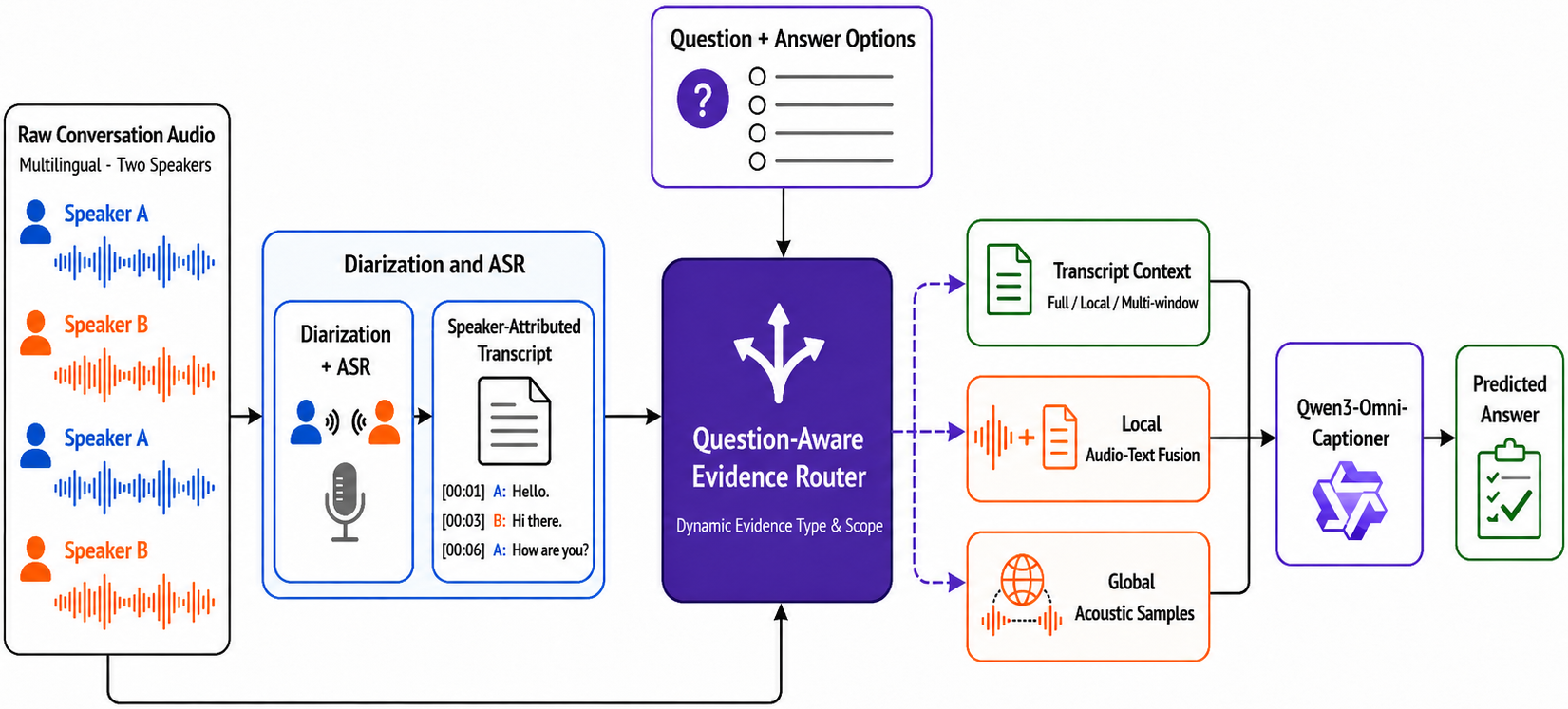}
  \caption{System overview for MLC-SLM Tasks 1 and 2. The front-end produces speaker-attributed transcripts, and the Task 2 router selects transcript, local audio-text, or global acoustic evidence before answer prediction.}
  \label{fig:overview}
\end{figure*}

Our submission is a single reproducible pipeline rather than a result-level ensemble. As shown in Fig.~\ref{fig:overview}, raw conversations are first converted into timestamped speaker-attributed transcripts. The Task 1 branch performs VAD, diarization, multilingual ASR, and STM normalization. The Task 2 branch then treats the STM as the textual backbone and uses the original waveform only when the routed question requires acoustic or speaker-sensitive evidence.

Fig.~\ref{fig:overview} highlights the two-stage design of the system. The diarization-ASR stage converts raw multilingual two-speaker audio into a structured transcript with speaker and time information. The evidence-routing stage combines this transcript, the waveform, and the question-answer pair to build a question-specific answer input.

\subsection{Diarization-ASR Front-end}

The diarization-ASR front-end provides speaker-attributed transcripts for both Task 1 evaluation and Task 2 evidence construction. It follows a cascade design, which remains the dominant Task 2 approach in the MLC-SLM summary~\cite{mu2025summary}. We use FSMN VAD to detect speech regions, RedimNet2 to extract speaker embeddings, and spectral clustering to estimate speaker turns. Subsegments use a 1.5 s duration and a 0.75 s step, with a 0.20 s minimum turn duration. After diarization, adjacent turns from the same speaker are concatenated up to 20 seconds when the inter-turn gap is at most 0.3 seconds. This post-processing is motivated by the same observation reported in previous systems: very short diarization segments deprive ASR models of local linguistic context and can increase tcpMER~\cite{xue2025teaaslp,mu2025summary}.

For Task 1, decoded hypotheses are sorted by timestamp and written in the official speaker-attributed STM format. Text normalization includes Unicode NFC normalization, invisible-character removal, lowercasing, punctuation removal, whitespace cleanup, and character-level tokenization for Japanese, Korean, and Thai. This front-end is important for Task 2 because the routing module relies on timestamps and speaker IDs to select local windows and speaker-linked evidence.

\subsection{Evidence Routing Formulation}

Let $X$ be the waveform, $T$ the diarized transcript, $q$ the question, and $C$ the answer options. The system constructs a structured routing label
\begin{equation}
  z=(e,s),
\end{equation}
where $e$ is the evidence type and $s$ is the context scope. In implementation, $z$ is obtained by combining coarse LLM predictions with deterministic timestamp and speaker-cue parsing. Given $z$, the system constructs routed evidence $E(q,C,X,T,z)$, and the answer model predicts
\begin{equation}
  \hat{c}=\arg\max_{c\in C}p(c\mid q,C,E).
\end{equation}

The evidence type dimension specifies what information is needed to answer the question. The scope dimension specifies where that information should be taken from. We found this two-dimensional label space to cover the dominant evidence patterns in the development set: questions typically ask about textual content, content plus acoustic realization, speaker-sensitive information, or global acoustic properties, with evidence concentrated in a local region, multiple regions, speaker-linked turns, or the whole conversation.

\noindent\textbf{Evidence type $e$.}
\begin{center}
\small
\setlength{\tabcolsep}{5pt}
\renewcommand{\arraystretch}{1.08}
\begin{tabularx}{0.98\linewidth}{L{0.34\linewidth}Y}
  \toprule
  Label & Routed evidence \\
  \midrule
  transcript-semantic & Semantic or discourse questions answered from full or local transcript context. \\
  semantic-acoustic & Content questions that require local acoustic cues; materialized as local transcript plus local audio. \\
  speaker-sensitive & Speaker attribution, timbre, or role questions; materialized from speaker-attributed transcript plus audio. \\
  global-acoustic & Anchorless acoustic questions; materialized as capped full transcript plus compact global audio. \\
  \bottomrule
\end{tabularx}
\end{center}

\noindent\textbf{Context scope $s$.}
\begin{center}
\small
\setlength{\tabcolsep}{5pt}
\renewcommand{\arraystretch}{1.08}
\begin{tabularx}{0.98\linewidth}{L{0.32\linewidth}Y}
  \toprule
  Scope & Trigger and construction \\
  \midrule
  local-window & Explicit timestamp or local event; extract transcript and/or audio around the anchor. \\
  multi-window & Multiple timestamps or comparison questions; extract several local windows. \\
  speaker-linking & Speaker attribution or who-said-what questions; collect speaker-linked turns and audio. \\
  global-compact & No anchor or whole-dialogue questions; keep long transcript context, capped when necessary, and attach sampled clips when acoustic. \\
  \bottomrule
\end{tabularx}
\end{center}

\subsection{Evidence Materialization}

Given $z=(e,s)$, the system materializes transcript and audio evidence. STM turns are first merged into chronological speaker-attributed blocks and serialized with timestamps and speaker IDs. For transcript-semantic questions with global-compact scope, the router keeps a capped full transcript to preserve discourse context. For local-window and multi-window scopes, it extracts transcript blocks overlapping each anchor in $[t_{\rm start}-12\,{\rm s},t_{\rm end}+12\,{\rm s}]$. This is a padded anchor span, rather than a fixed 24-s window. For speaker-linking scope, deterministic parsing also extracts timestamp expressions from the answer options, allowing option-local references to contribute evidence.

Audio evidence is attached for semantic-acoustic, speaker-sensitive, and global-acoustic labels. For timestamped semantic-acoustic or speaker-sensitive questions, the system uses local audio-text fusion: each routed span is padded by 0.5 s on both sides, capped at 30 s, and paired with the wider transcript window above. For anchorless global-acoustic questions in M5--M6, the system keeps the capped full transcript and attaches compact global audio. Timestamped STM blocks form the audio candidate pool; we sample without replacement using a fixed pseudorandom seed, retain at most 12 blocks, cap each clip at 30 s, and restore the selected clips to chronological order. If no timed STM block is available, the fallback samples uniformly along the recording timeline. The clips remain separate \texttt{audio\_url} blocks in one model request; they are neither concatenated nor answered by voting.

This design makes the router question-aware but not content-searching: it exploits explicit timestamps, option-local references, and speaker cues, but does not retrieve implicit evidence spans from the transcript or waveform.

\subsection{Hybrid Router Implementation}

The hybrid router is implemented in three steps. First, a Qwen3-Omni-30B-A3B-Captioner classifier predicts coarse evidence type and scope from only the question and answer options, without access to the transcript, waveform, gold answer, or correctness feedback. Second, deterministic parsing extracts timestamps and speaker/role cues from the question and options. Third, the coarse labels, anchors, and lexical cues are jointly mapped to executable routing labels: global semantic questions use transcript-semantic/global-compact evidence; local semantic questions use transcript-semantic/local-window evidence; local acoustic questions use semantic-acoustic/local-window or speaker-sensitive/speaker-linking evidence; and global acoustic questions use global-acoustic/global-compact evidence. Invalid classifier outputs fall back to deterministic routing.

\subsection{Task 2 System Variants}

The ablation systems are grouped by evidence policy. M1--M3 isolate transcript routing: M1 uses the full diarized transcript for every question, M2 uses regex/time-anchor transcript routing, and M3 uses offline-label transcript routing. These variants do not provide waveform audio to the answer model. M4--M6 progressively add the proposed audio-text routing. M4 adds local audio-text fusion for timestamped semantic-acoustic and speaker-sensitive questions. M5 adds compact global-acoustic sampling for acoustic or speaker questions without explicit timestamps. M6 replaces offline labels with the hybrid router.

The audio-centric diagnostic systems in Table~\ref{tab:audio_only} are not the final method. They keep the question and answer options as text, intentionally avoid using ASR transcripts as conversation evidence, and vary the requested global sampled-clip budget. Their purpose is to test whether increasing waveform coverage can replace transcript-backed routing.

\subsection{Training and Adaptation}

For Task 1, supervised training and adaptation use the official timestamps, speaker labels, and transcripts. Following the trend in recent MLC-SLM systems toward parameter-efficient adaptation and staged training~\cite{mu2025summary,gao2025triplex,meng2025ilt}, we fine-tune Qwen3-ASR 1.7B for two epochs with LoRA on the official MLC-SLM training data. Recent LoRA extensions improve parameter utilization through dense low-rank adaptation~\cite{mu-etal-2025-denselora} and incorporate graph structure into low-rank updates for recommendation~\cite{mu-etal-2026-graphlora}, illustrating the broader flexibility of parameter-efficient adaptation. The goal of our LoRA setup is to adapt the recognizer to multilingual conversational speech while preserving the general recognition ability of the pretrained model.

\begin{table}[t]
  \caption{External data and pretrained components used for diarization, ASR, routing, and QA.}
  \label{tab:resources}
  \centering
  \begingroup
  \small
  \setlength{\tabcolsep}{5pt}
  \renewcommand{\arraystretch}{1.08}
  \begin{tabularx}{0.98\linewidth}{L{0.48\linewidth}Y}
    \toprule
    Resource & Role in the system \\
    \midrule
    Official MLC-SLM data~\cite{mlcslm2026,mu2025summary} & ASR fine-tuning and evidence-routing design. \\
    FSMN VAD~\cite{gao2023funasr} & Speech activity detection. \\
    RedimNet2~\cite{yakovlev2026redimnet2} & Speaker embedding extraction. \\
    Qwen3-ASR 1.7B~\cite{shi2026qwen3asr} & Segment transcription. \\
    Qwen3-Omni Captioner~\cite{xu2025qwen3omni} & QA inference and context labeling. \\
    \bottomrule
  \end{tabularx}
  \endgroup
\end{table}

For Task 2, we do not train or fine-tune Qwen3-Omni for either routing or answer prediction. System development concerns evidence selection rather than additional model adaptation. Regex labels are deterministic and inexpensive, but only capture explicit timestamp and keyword patterns. Offline labels provide structured labels for controlled development-set ablations. Dynamic labels are generated online by an LLM question analyzer and are used in the final M6 system. The dynamic labeler only observes the question and answer options; it does not use gold answers, analysis buckets, or correctness feedback.

\begin{table}[t]
  \caption{Task 1 tcpMER (\%) on development and evaluation sets.}
  \label{tab:task1_results}
  \centering
  \small
  \setlength{\tabcolsep}{6pt}
  \begin{tabular*}{0.78\linewidth}{l@{\extracolsep{\fill}}cc}
    \toprule
    System & Dev & Eval \\
    \midrule
    Official baseline & 79.15 & -- \\
    Ours & 25.70 & 18.44 \\
    \bottomrule
  \end{tabular*}
\end{table}

\section{Experiments and Analysis}

\subsection{Datasets, Metrics, and Setup}

The MLC-SLM 2026 training set contains about 2100 hours of multilingual two-speaker conversational speech. We use the development set for system design and ablation; its Task 2 subset contains 4,500 multiple-choice questions. The evaluation set is held out from training and tuning.

\nextleftpage
\sbox{\resulttables}{%
\begin{minipage}{\textwidth}

\begin{center}
  \captionof{table}{Task 2 development accuracy (\%) for official baseline and transcript-backed routing systems. Delta columns are absolute points.}
  \label{tab:main_results}
  \small
  \setlength{\tabcolsep}{2.8pt}
  \begin{tabular}{L{0.04\textwidth}L{0.213\textwidth}C{0.073\textwidth}C{0.073\textwidth}C{0.073\textwidth}C{0.088\textwidth}C{0.104\textwidth}C{0.105\textwidth}C{0.105\textwidth}}
    \toprule
    ID & System & Overall & \shortstack{$\Delta$ vs.\\M1} & \shortstack{$\Delta$ vs.\\prev.} & No-anchor & Timestamped & \shortstack{No-anchor\\acoustic} & \shortstack{No-anchor\\semantic} \\
    \midrule
    O0 & Official baseline & 35.33 & -- & -- & -- & -- & -- & -- \\
    M1 & Full transcript & 93.16 & $+0.00$ & -- & 94.10 & 92.66 & 92.08 & 94.39 \\
    M2 & Regex text routing & 92.89 & $-0.27$ & $-0.27$ & 93.99 & 92.27 & 90.83 & 94.45 \\
    M3 & Offline-label text routing & 92.69 & $-0.47$ & $-0.20$ & 93.52 & 92.23 & 90.42 & 93.97 \\
    M4 & M3 + local audio-text fusion & 94.22 & $+1.06$ & $+1.53$ & 95.63 & 93.36 & 95.42 & 95.66 \\
    M5 & M4 + global audio & 94.40 & $+1.24$ & $+0.18$ & 95.84 & 93.51 & \textbf{96.67} & 95.72 \\
    M6 & M5 + hybrid router & \textbf{94.84} & $\mathbf{+1.68}$ & $\mathbf{+0.44}$ & \textbf{96.31} & \textbf{93.90} & 95.83 & \textbf{96.38} \\
    \bottomrule
  \end{tabular}
\end{center}

\begin{center}
  \captionof{table}{Task 2 audio-centric diagnostic accuracy (\%) by requested clip budget.}
  \label{tab:audio_only}
  \small
  \setlength{\tabcolsep}{2.8pt}
  \begin{tabular}{L{0.04\textwidth}L{0.25\textwidth}C{0.06\textwidth}C{0.08\textwidth}C{0.09\textwidth}C{0.105\textwidth}C{0.105\textwidth}C{0.105\textwidth}}
    \toprule
    ID & Routing & Clips & Overall & No-anchor & Timestamped & \shortstack{No-anchor\\acoustic} & \shortstack{No-anchor\\semantic} \\
    \midrule
    B1 & Global audio only & 10 & 84.71 & 80.61 & 87.88 & 77.08 & 81.12 \\
    B2 & Global audio only & 50 & \textbf{92.07} & 92.52 & \textbf{91.96} & 83.33 & 93.85 \\
    B3 & Regex-routed audio only & 10 & 86.62 & 80.87 & 91.03 & 79.17 & 81.12 \\
    B4 & Regex-routed audio only & 50 & 91.96 & \textbf{92.78} & 91.53 & \textbf{84.58} & \textbf{93.97} \\
    B5 & Label-routed audio only & 10 & 87.11 & 81.03 & 91.69 & 77.08 & 81.60 \\
    B6 & Label-routed audio only & 50 & 91.31 & 91.20 & 91.53 & 82.92 & 92.40 \\
    \bottomrule
  \end{tabular}
\end{center}

\setlength{\multicolsep}{4pt}
\begin{multicols}{2}
\subsection{Task 1 Results}

Table~\ref{tab:task1_results} reports Task 1 tcpMER results. The official development baseline is 79.15\%, while our cascaded diarization-ASR system obtains 25.70\% on the development set and 18.44\% on the evaluation set. Because the development and evaluation splits differ, these numbers should not be read as a controlled split-to-split comparison. They mainly show that the transcripts used by Task 2 are strong enough to form a competitive text baseline, but still imperfect; therefore, acoustic evidence is most likely to help when the question depends on cues not represented in the transcript or when ASR makes acoustically plausible errors.

\subsection{Main Task 2 Ablation}

Table~\ref{tab:main_results} shows that the full-transcript baseline is already strong at 93.16\%. Static transcript routing does not improve it: M2 and M3 decrease by 0.27 and 0.47 points, suggesting that context reduction can remove useful discourse information. Local audio-text fusion gives the largest single-step gain: M4 improves over M3 by 1.53 points and over M1 by 1.06 points. Adding global audio in M5 yields a further 0.18-point gain and raises no-anchor acoustic accuracy from 95.42\% to 96.67\%.

Replacing offline labels with dynamic labels provides a further 0.44-point gain: M6 reaches 94.84\%, improving over M1 by 1.68 points (about 76 questions). Relative to M5, M6 improves no-anchor semantic accuracy by 0.66 points and timestamped accuracy by 0.39 points, although no-anchor acoustic accuracy decreases by 0.84 points. This pattern suggests that dynamic routing improves question-conditioned evidence allocation rather than merely activating more audio. However, predicted labels jointly determine scope and modality; without a matched dynamic-text control, M5--M6 cannot isolate label quality from the induced evidence allocation. The lack of paired significance tests further precludes strict causal attribution.

Fine-grained offline buckets show the same trend. Compared with M1, M6 improves no-anchor/global questions from 94.10\% to 96.31\% and timestamped/local questions from 92.66\% to 93.90\%. Within no-anchor questions, both acoustic and semantic categories improve, from 92.08\% to 95.83\% and from 94.39\% to 96.38\%, respectively. Since these buckets are used only for analysis and some subsets are small, the results are treated as supporting evidence rather than standalone claims.

\subsection{Audio-budget Clip-count and Routing Analysis}

Table~\ref{tab:audio_only} analyzes audio-centric evidence under different requested sampled-clip budgets and routing strategies. Across all three routing modes, increasing the requested budget from 10 to 50 clips improves accuracy, indicating that very sparse audio sampling may under-cover relevant evidence. However, the best audio-centric diagnostic configuration, B2 at 92.07\%, remains below both the full-transcript baseline and the final routing system. This supports question-conditioned transcript-backed routing over fixed audio-centric inputs.

\section{Conclusions}

We presented the Bairong system for the MLC-SLM 2026 Challenge. Our Task 1 diarization-ASR cascade obtains 25.70\% and 18.44\% tcpMER on the development and evaluation sets. For Task 2, relative to the offline-label text-routing control, local audio-text fusion gives the largest step gain ($+1.53$) and improves over the full-transcript baseline by 1.06 points. Replacing offline labels with the hybrid question-aware router adds a further 0.44 points and yields the best overall result of 94.84\%. Although the ablation does not fully separate label quality from its induced evidence allocation, the results support hybrid question-aware routing: transcripts remain the backbone, audio supplies complementary acoustic evidence, and LLM estimates combined with deterministic cues refine which evidence is exposed for each question.

\section{Acknowledgements}

We thank the MLC-SLM 2026 organizers for providing the multilingual conversational speech dataset, baseline systems, and evaluation platform.
\end{multicols}

\end{minipage}
}
\noindent\makebox[0pt][l]{\usebox{\resulttables}}
\newpage
\null

\end{document}